# Performance Analysis of YOLOv11 and YOLOv8 for Mixed Traffic Object Detection under Adverse Weather Conditions in Developing Countries

Quoc Thuan Nguyen[1], Ha Anh Vu [2], Ngo Dang Thanh Ngan[3] and Minh Phuc Hoang Ngoc[4]

[1] FPT University, Ho Chi Minh City, Vietnam
Thuannguyenquoc.work@gmail.com
[2] FPT University, Ho Chi Minh City, Vietnam
AnhVH54@fe.edu.vn
[3] FPT University, Ho Chi Minh City, Vietnam
NganNDT2@fe.edu.vn
[4] FPT University, Ho Chi Minh City, Vietnam
Hoangngocminhphuc.work@gmail.com

**Abstract.** In modern vehicular systems, robust performance under harsh conditions has become a critical problem of autonomous driving. Our study delivers a comprehensive evaluation of the newest iteration of the YOLO series, which is YOLOv11 Nano architecture benchmarked against the widely adopted YOLOv8 Nano as a baseline on a custom fused dataset that combines the Indian Driving Dataset (IDD) [1] and Berkeley Deep Drive Dataset (BDD100K) [2]. We have analyzed the trade-offs among detection accuracy, inference speed, and computational efficiency in high-entropy scenarios involving dense mixed traffic, rain, and low-light conditions. Specifically, YOLOv11n achieves a mean Average Precision (mAP@50) of 46.6%, with a notable 3.2% improvement in Precision over the baseline, effectively reducing false positives in cluttered scenes. Furthermore, the proposed model exhibits enhanced energy efficiency, requiring 22% fewer FLOPs (6.3G vs. 8.1G) while maintaining real-time inference speed of 70.9 FPS on a Tesla T4 GPU, offering an optimal trade-off for safety-critical edge deployment.

## 1 Introduction

Intelligent Transportation Systems (ITS) have become an irreplaceable component of safe driving and improve mobility efficiency. However, most existing object detection research focuses on well-structured traffic conditions in developed countries, typically using datasets such as COCO or KITTI. In developing countries (such as Vietnam, India, etc.), traffic conditions are way more complicated, where diverse vehicles such as cars, motorcycles, buses, and indigenous vehicles (e.g., auto-rickshaws, motorbikes) share the same lane [1]. Additionally, the weather conditions in these countries are relatively harsh, as you will constantly encounter adverse conditions such as tropical rain or dense fog, which degrade visibility and significantly affect the robustness of most object detection models.

To address these problems, we present a comprehensive evaluation of the state-of-the-art YOLOv11 architecture for mixed-traffic object detection. We apply a Data Fusion strategy that integrates the unique vehicle classes from IDD with the harsh environmental samples from BDD100K [2]. By training the model on this hybrid dataset, we aim to create a detector that is both "culturally adaptive" (recognizing local vehicles) and "environmentally robust" (seeing through rain/night).

The main contributions of this paper are characterized by an empirical performance analysis of YOLOv11 compared to its predecessor, YOLOv8, focusing on inference speed (FPS) and accuracy (mAP). Futhermore, we build up a Hybrid dataset, which will help increase model robustness and prove that the proposed training strategy will produce a well-operated model in reality.

Note that the first paragraph of a section or subsection is not indented. The first paragraphs that follow a table, figure, equation, etc., do not have an indent, either.

## 2 Methodology

### 2.1 Dataset preparation

To ensure the robustness of this model under mixed-traffic conditions and adverse weather, we developed a strategy that combines the strengths of two datasets: IDD (specializing in indigenous vehicle types typical of developing countries) [1] and BDD100K(specializing in extreme weather scenarios) [2], which then result in a final training set comprising a total of 8,674 images. This data preparation process included three main stages:

**Phase 1:** Domain-Specific Data Selection. The majority of the data (7,981 images, approximately 92% of the dataset) is sourced from the IDD dataset [1]. This component captures the chaotic nature of unstructured traffic, including high-density classes such as "Motorcycle" and "Rider", as well as region-specific vehicles like "Autorickshaw". This aligns directly with our research objective of contributing to autonomous vehicle solutions for countries like Vietnam, India, and Indonesia, a challenge recently highlighted by Min et al. [3] and Wibowo et al. [4]. In addition, we combined the BDD100K dataset [2] to provide the model with essential environmental information under unpredictable and harsh weather conditions, a critical variable that the IDD dataset lacks [2].

**Phase 2:** Attribute-Based Filtering. Instead of using all images from the BDD100K dataset, which would lead to an imbalance between vehicle classes and bias towards standard vehicles. To prevent this, we implemented a strict attribute-based filtering mechanism. From the BDD100K dataset, we selected only images with attributes describing harsh weather conditions, such as “Rainy”, “Foggy”, and “Night”. This process resulted in a refined set of 693 sample images. These sample images will serve as "hard negatives" during training, forcing the model to learn to operate effectively in harsh weather conditions and addressing the limitations of the standard detectors reviewed by Wei et al. [5]. For the IDD dataset, we only filtered and retained images containing 8 classes: Car, Bus, Truck, Motorcycle, Rider, Person, Auto-rickshaw, and Animal.
**Phase 3:** Label Harmonization and Standardization. Since the two datasets utilize different annotation formats, direct training on YOLOv11 was not possible. Therefore, the data must be synchronized into a unified schema consisting of the 8 classes mentioned above. The dataset follows a stratified split strategy (Training: 70%, Validation: 20%, Testing: 10%) to maintain class distribution consistency. Crucially, to address potential data leakage, the split was performed based on unique video sequence IDs provided in the metadata. This ensures that frames from the same driving sequence do not appear in both the training and test sets, guaranteeing a fair evaluation of the model's generalization capability. Notably, the 'Auto-rickshaw' class was prioritized to preserve the local context. Finally, all annotations were converted to the standard YOLO format (normalized x, y, w, h coordinates) as described in the YOLO framework review [6], and images were resized to **640x640** pixels for optimal training efficiency. The detailed statistics of the training set are presented in Table 1. The distribution reflects a significant long-tail challenge: the "Animal" class accounts for only 0.68% of the total instances (1230 annotations), compared to the dominant "Car" and "Motorcycle" classes.

**Table 1.** Composition and Class Distribution of the Training Set.

| Class | IDD source | BDD source | Total instances |
|---|---|---|---|
| Car | 32288 | 5903 | 38191 |
| Bus | 6307 | 120 | 6427 |
| Truck | 9034 | 193 | 9227 |
| Motorcycle | 37629 | 110 | 37739 |
| Rider | 38050 | 39 | 38089 |
| Person | 35169 | 1358 | 36527 |
| Autorickshaw | 13329 | 0 | 13329 |
| Animal | 1230 | 0 | 1230 |

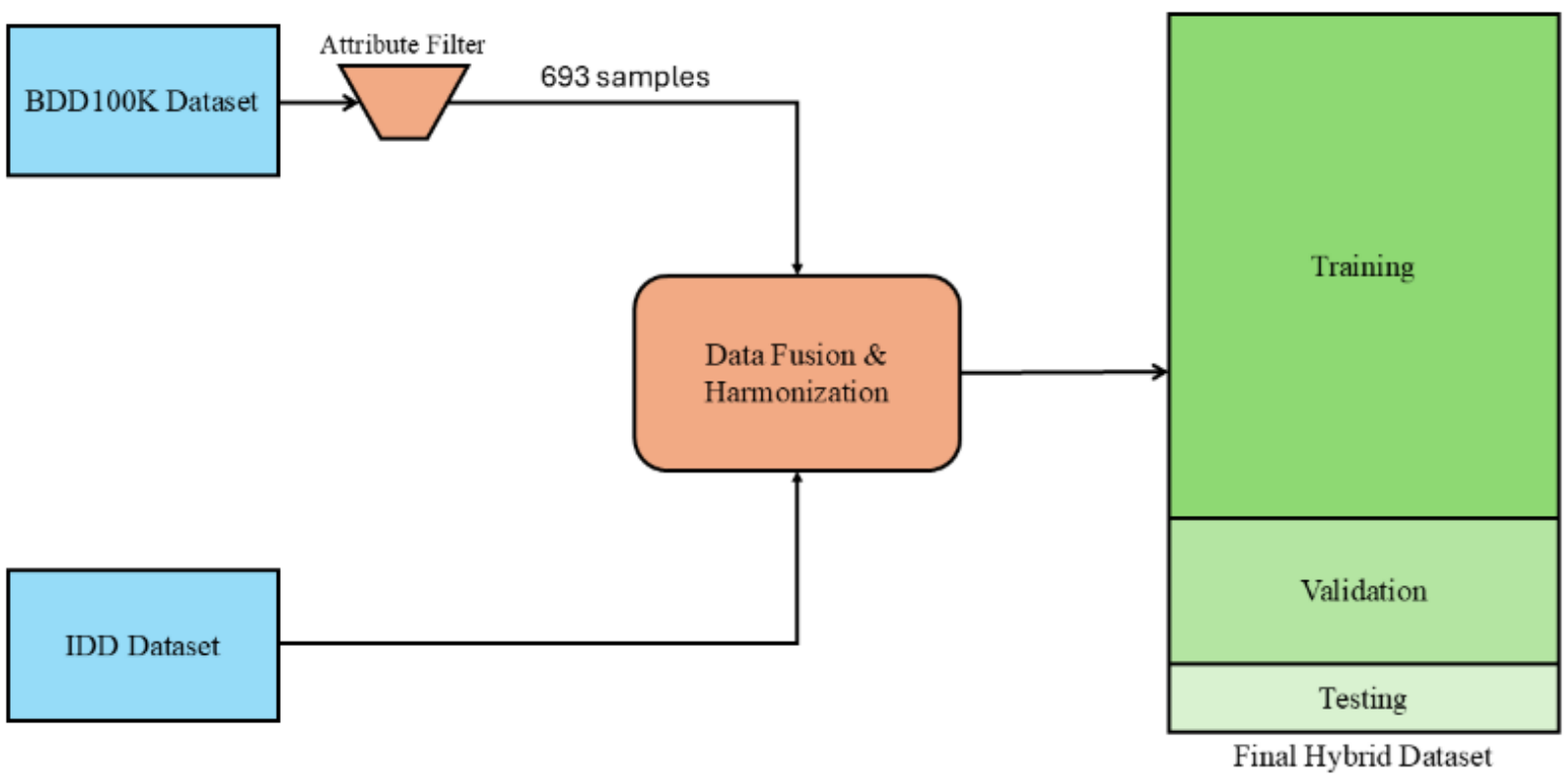

**Fig. 1.** The data preparation pipeline.

### 2.2 YOLOv8 Architecture

The deployment of YOLOv8 as our baseline is due to its status as an industry standard for balancing latency and accuracy [5]. Its architecture improves upon predecessors by introducing the C2f module (Fig. 2), which enriches gradient flow for better feature representation without increasing computational load [7]. Additionally, YOLOv8 utilizes an anchor-free, decoupled head; this design handles the irregular aspect ratios of vehicles like autorickshaws by predicting object centers directly, while simultaneously separating classification and regression tasks to improve training convergence [6].

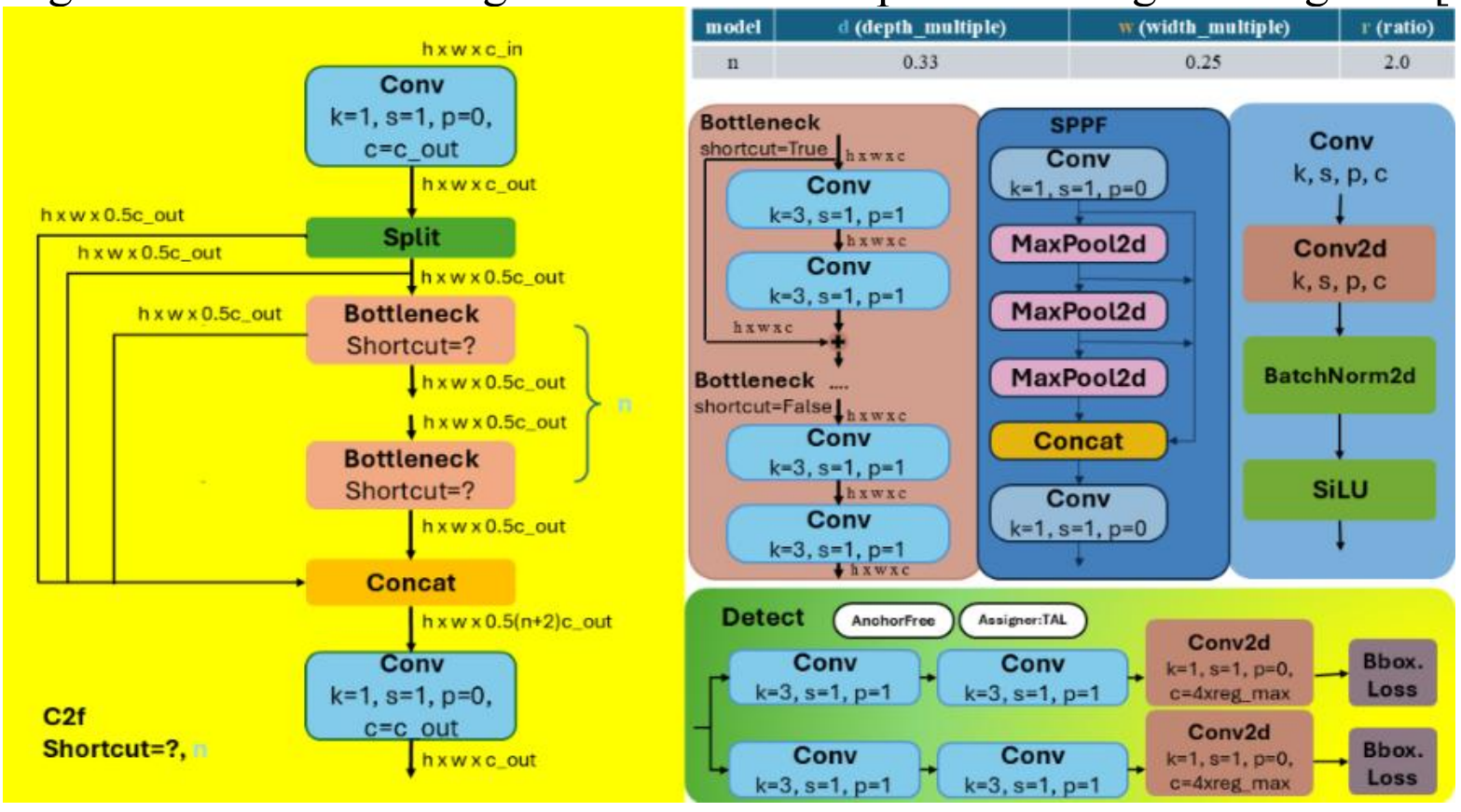

**Fig. 2.** The architecture of the YOLOv8 model [7].

### 2.3 YOLOv11 Architecture

We selected YOLOv11, the latest Ultralytics iteration, for its optimal balance between inference speed and detection accuracy, a critical requirement for autonomous driving systems [5, 6]. This model introduces specific architectural enhancements over previous versions, notably replacing the C2f module with the C3k2 block to improve semantic feature extraction in complex backgrounds. This modification enables better discrimination between similar objects, such as riders and motorcycles, even in cluttered visual environments. Furthermore, integrating a C2PSA (Cross-Stage Partial with Spatial Attention) module at the end of the backbone enables the network to prioritize informative regions while suppressing environmental noise, such as rain or low-light artifacts, making YOLOv11 ideal for real-time edge deployment [8].

To evaluate these architectural shifts, we use YOLOv8 as a comparative baseline due to its established stability and status as the industry standard for real-time detection [5, 6]. Benchmarking directly against this predecessor allows us to isolate and quantify the specific performance gains attributed to the transition from C2f to C3k2 blocks [9]. Consequently, this comparison verifies whether YOLOv11's theoretical enhancements translate into practical improvements for detecting "hard" classes, such as autorickshaws, under visibility degradation, rather than offering solely computational reductions.

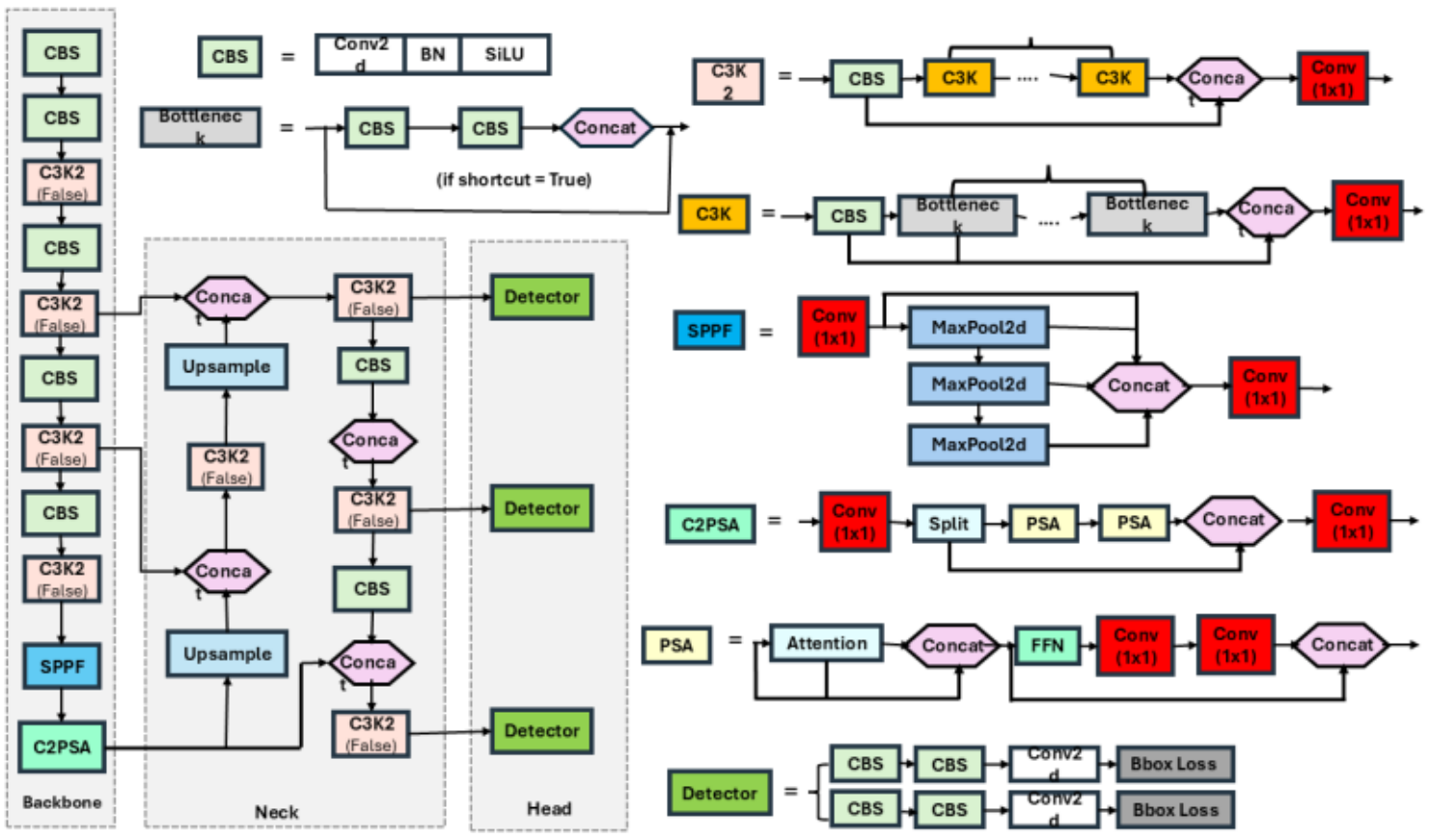

**Fig. 3.** The architecture of the YOLOv11 model [9].

## 2.4 Experimental Setup

To ensure strict evaluation of the YOLOv11n and YOLOv8n models, we configured both as "Nano" variants, specifically optimized for edge computing constraints. To secure experimental validity and reproducibility, all experiments were conducted in a standardized cloud-based environment using Google Colab, featuring an NVIDIA Tesla T4 GPU (16GB VRAM), PyTorch 2.0 [10], and the Ultralytics library [11]. A critical aspect of this study is to isolate the performance gains attributable solely to architectural advancements (such as the C3k2 block) rather than to hyperparameter tuning. Therefore, we kept the standard default hyperparameters provided by Ultralytics for both models. By keeping the training configuration similar, we can isolate the specific contributions of the architectural improvements to the overall performance.

Regarding the training strategy, we employed Transfer Learning to mitigate overfitting risk given the limited size of the adverse-weather subset (906 samples). Instead of training from scratch, both models were initialized with weights pre-trained on the COCO dataset [12], thereby inheriting robust feature-extraction capabilities.
The configuration for both experiments is configured as follows:

- Optimizer: Auto-configured (defaulting to SGD with a momentum of 0.937).
- Learning Rate: Initial lr_0 = 0.01 with a linear decay strategy to ensure stable convergence.
- Epochs: 50, with an early stopping mechanism (patience = 10 epochs) to prevent unnecessary computation if the model performance plateaus.
- Batch Size: 16.
- Input Resolution: All images were resized to 640 x 640 pixels.

### 2.5 Performance Metrics

We evaluated YOLOv11n and YOLOv8n for autonomous driving in adverse weather using standard metrics [13]: Precision, Recall, Mean Average Precision (mAP), and Inference Speed.
***Precision (P):***
Precision measures the accuracy of positive predictions. High precision is critical to minimize false alarms and prevent unnecessary emergency braking. It is defined as the ratio of true positives (TP) to all positive detections (TP + FP):

$$P = \frac{TP}{TP+FP} \tag{1}$$

***Recall (R):***
Recall (sensitivity) measures the model's ability to identify all actual positive instances. High recall is vital for safety in autonomous driving, ensuring that road users are detected and that potential hazards (False Negatives) are minimized. It is expressed as:

$$R = \frac{TP}{TP + FN} \tag{2}$$

where FN represents the number of False Negatives (missed detections).
***Mean Average Precision (mAP@0.5):*** Mean Average Precision (mAP) evaluates the trade-off between Precision and Recall. We report mAP@0.5 (IoU threshold of 0.5), where higher scores indicate better overall detection performance [13]. It is computed by averaging the Average Precision (AP) across all N classes:

$$mAP = \frac{1}{N} \sum_{i=1}^{N} AP_i \tag{3}$$

***Inference Speed (FPS)***
Beyond accuracy, real-time processing capability is a requirement for a model to be deployed on edge devices. We evaluate the computational efficiency using Frames Per Second (FPS) measured on the NVIDIA Tesla T4 GPU. As highlighted in recent comprehensive reviews of YOLO algorithms [5], evaluating FPS is essential to verify whether the model meets the stringent latency requirements for timely decision-making in mixed traffic environments.

## 3 Experiments and Results

### 3.1 Training Dynamics and Convergence Analysis.

The training progression of the YOLOv11n model compared to the YOLOv8n is presented in Fig. 3, which illustrates the validation Mean Average Precision (mAP@0.5) and Training Box Loss over 50 epochs.
As shown in Fig. 3(a), the YOLOv11n architecture (solid red line) converges faster than the baseline. Specifically, the YOLOv11n model achieves rapid learning stabilization within the first 20 epochs, while YOLOv8n shows slower adaptation. In terms of detection accuracy, YOLOv11n consistently maintains a high mAP@0.5, reaching 0.4661, a slight improvement over YOLOv8n's 0.4611. This gap validates the effectiveness of the C3k2 blocks in extracting robust features from the fused dataset, especially under the adverse weather subset.

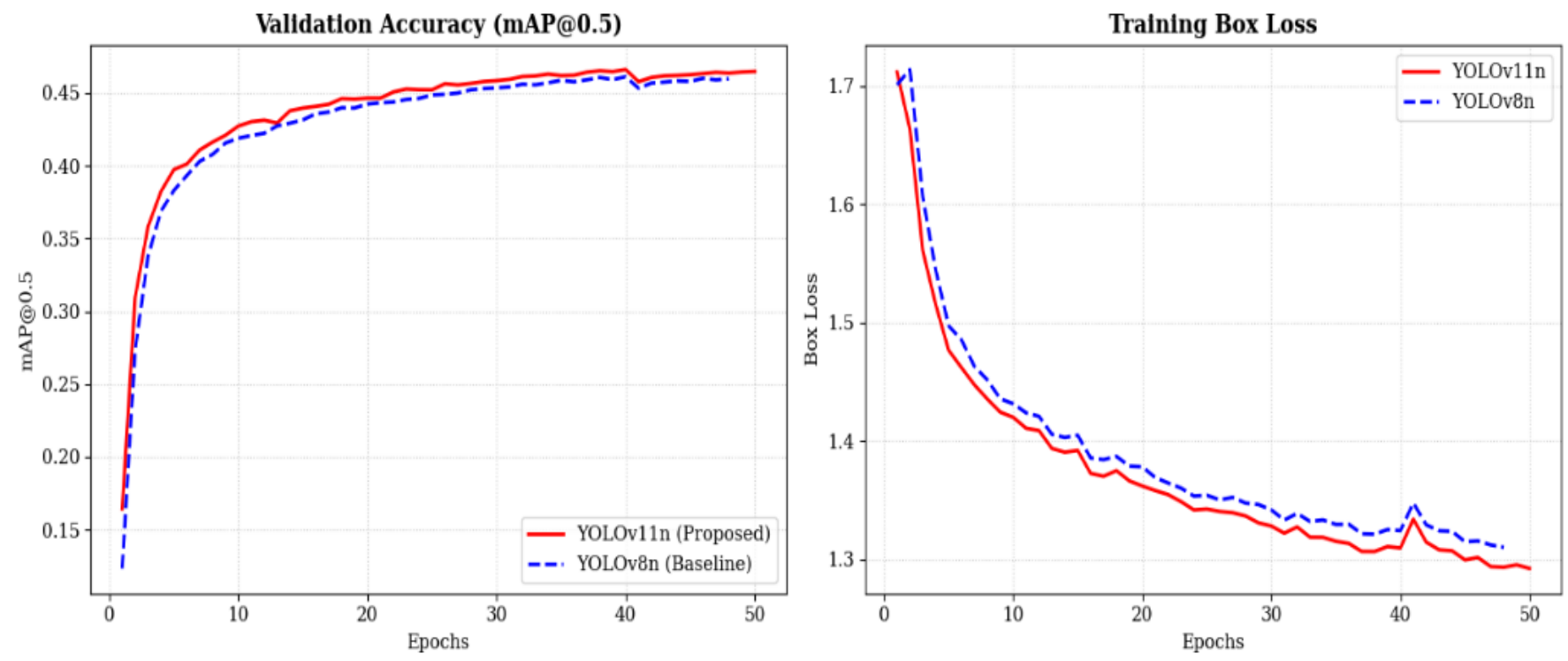

**Fig. 4.** Comparative training dynamics between YOLOv11n and YOLOv8n over 50 epochs: Validation mAP@0.5 curves (a), Training Box Loss (b).

Furthermore, the Box Loss analysis in Fig. 3 (b) confirms these findings. The YOLOv11n model shows a sharp drop in loss values during the initial phase and converges to a low steady-state error of about 1.2922. This indicates that the architectural enhancements in YOLOv11 enable more accurate bounding-box regression, which is essential for localizing objects in unstructured traffic scenarios.

### 3.2 Quantitative Evaluation

Evaluating the trade-off between accuracy and efficiency involves benchmarking the YOLOv11n model against YOLOv8n on an NVIDIA Tesla T4 GPU. The results are shown in Table 2, where the YOLOv11n architecture achieves a delicate balance between model complexity and detection accuracy.

The most notable advancement is that YOLOv11n precision achieves 0.686, outperforming the YOLOv8n precision (0.654) by 3.2%. This enhancement is critical to the safety of autonomous driving. Additionally, the model maintains a consistent gain in mAP@0.5 (+0.5%), illustrating robust feature extraction capabilities despite the challenging environment.
Furthermore, the YOLOv11n architecture is proven to be more lightweight, requiring 14% fewer parameters (2.58M) and 22% fewer FLOPs (6.3G) than YOLOv8n. This reduction in computational load makes it highly suitable for storage-constrained edge devices. However, on the Tesla T4 GPU, YOLOv8n exhibits lower latency (10.7 ms) and a higher throughput of 93.4 FPS, compared to 70.9 FPS for YOLOv11n. This suggests that while YOLOv11n reduces the computational volume (FLOPs), its architectural design (C3k2 blocks) may introduce a slightly higher operational overhead per layer than the highly optimized C2f blocks in YOLOv8n. It is observed that while YOLOv11n reduces FLOPs by 22% compared to YOLOv8n, its inference speed is lower (70.9 FPS vs. 93.4 FPS). This discrepancy aligns with the findings in ShuffleNet V2 [14], which demonstrated that FLOPs is an indirect metric and does not strictly correlate with latency due to Memory Access Cost (MAC). YOLOv11n incorporates Depthwise Separable Convolutions in its C3k2 blocks and introduces C2PSA attention modules. While mathematically lighter, these fragmented operations incur higher MACs and are less parallelizable on the NVIDIA Tesla T4 architecture than the highly optimized dense convolutions (C2f) in YOLOv8, a phenomenon similarly observed in RepVGG [15]. However, for battery-powered edge devices where thermal throttling is governed by power consumption (closely related to FLOPs), YOLOv11n remains a theoretically ideal candidate for energy efficiency.

**Table 2.** Quantitative performance comparison between YOLOv8n and YOLOv11n on the mixed-traffic, adverse-weather dataset.

| Model | Parameters (M) | FLOP (G) | Precision (P) | Recall (R) | mAP@0.5 | Inference Time (ms) | FPS |
|---|---|---|---|---|---|---|---|
| YOLOv8n | 3.01 | 8.1 | 0.654 | 0.402 | 0.461 | 10.7 | 93.4 |
| YOLOv11n | 2.58 | 6.3 | 0.686 | 0.401 | 0.466 | 14.1 | 70.9 |

### 3.3 Quantitative Analysis

We have conducted a visual analysis of detection results on the test set to complement the quantitative metrics presented in Table 2. This qualitative assessment demonstrates how the architectural differences between YOLOv11n and YOLOv8n translate into observable behavioral changes in unstructured environments. One of the most critical challenges in the traffic environment of developing countries is the high degree of occlusion, where motorcyclists, autorickshaws, and pedestrians frequently move in tight clusters. As shown in Fig. 4, the proposed YOLOv11n demonstrates excellent ability in delineating adjacent objects. In scenarios where the baseline YOLOv8n merges two overlapping riders into a single bounding box or misses a partially occluded vehicle, YOLOv11n generates distinct and tight bounding boxes for each entity. This visual evidence corroborates the higher Precision (+3.2%) observed quantitatively, suggesting that the C3k2 blocks in YOLOv11 are more effective at extracting fine-grained features and suppressing background noise in cluttered scenes.

Despite the architectural advancements, both models exhibit similar failures in detecting minority classes. As you can see in Fig. 4, there is a failure case involving the "Animal" class under low-light conditions. Both YOLOv8n and YOLOv11n fail to detect the stray animal, misclassifying it as background. This behavior aligns with the extremely low Recall scores recorded for this specific class in our confusion matrices. This consistent limitation across both architectures indicates that the issue is primarily driven by data imbalance (insufficient training samples for animals in the adverse weather subset) rather than architectural deficiencies. This finding highlights the necessity for targeted data augmentation strategies in future work.

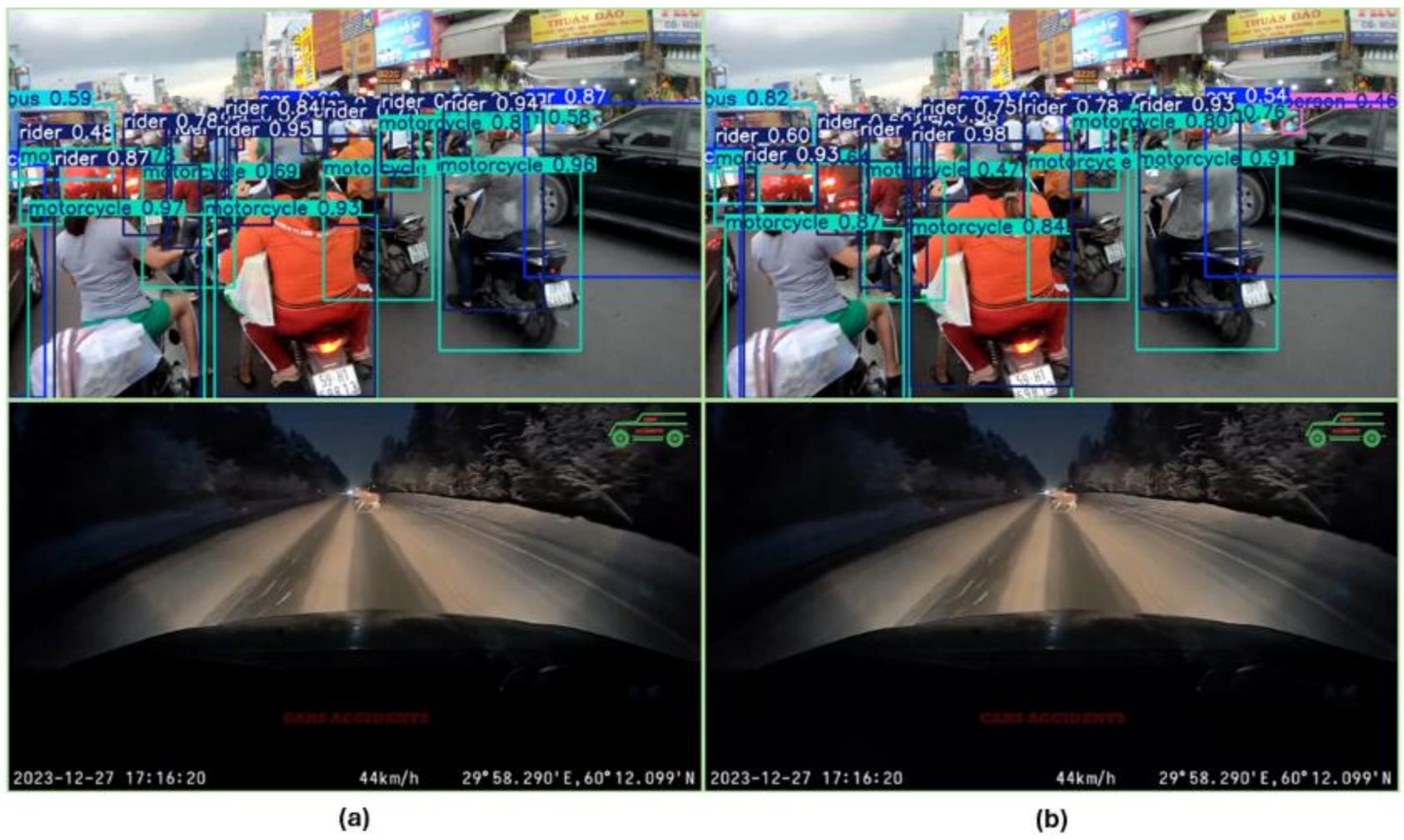

**Fig. 5.** Qualitative comparison of detection results of Yolov8n (a), Yolov11n (b).

### 3.4 Comparison with State-of-the-Art Methods

To position our proposed YOLOv11n within the broader research landscape, we conducted a comparative analysis against recent state-of-the-art (SOTA) studies focusing on unstructured traffic and adverse weather conditions. The quantitative results are summarized in Table 2. It is evident that our proposed model significantly surpasses standard baselines on unstructured benchmarks, such as Faster R-CNN on IDD [1] (+19.4% mAP). More importantly, in the context of adverse weather, YOLOv11n achieves higher accuracy (46.6%) than specialized adaptive frameworks such as IA-YOLO [16] (43.2%) and older architectures such as YOLOv3 [17] (38.5%). This confirms that the architectural integration of C3k2 blocks provides robust feature extraction capabilities that can handle environmental degradation (rain/night) without requiring complex image-preprocessing. We also acknowledge that studies using custom, localized datasets [4] or massive-scale benchmarks such as DriveIndia [18] report significantly higher mAP scores (up to 78.7%). However, these results must be contextualized. The performance in [18] is achieved using a training set of over 66,000 images and likely employs high-capacity models (Medium/Large variants), which typically operate below 40 FPS on standard GPUs. Similarly, the YOLOv8-Nano implementation on Indian roads by Padia et al. [19] reports a high Precision (60%) but suffers from a severe Recall drop (0.50), indicating a tendency to miss objects in complex scenes. In contrast to the previous approaches, our YOLOv11n offers a critical balance optimized for edge deployment. While our mAP is lower than that of large-scale models, we maintain a fine-grained inference speed of 70.9 FPS on the Tesla T4, significantly faster than the adaptive method in [16] (52 FPS) and the heavy baseline in [1]. This demonstrates that our proposed method is a considerable candidate for real-time applications where maintaining a high frame rate and energy efficiency is as critical as detection accuracy.

**Table 3.** Performance comparison with state-of-the-art methods on unstructured traffic and adverse weather datasets.

| Reference | Year | Method | Dataset Context | Metric (%) | FPS |
|---|---|---|---|---|---|
| [1] | 2019 | Faster R-CNN | IDD (Unstructured) | 27.2 (mAP) | 7 |
| [4] | 2023 | YOLOv7MOD | Mixed Traffic (Custom) | 94.6 (mAP) | 45 |
| [16] | 2022 | IA-YOLO | Adverse Weather | 43.2 (mAP) | 52 |
| [19] | 2024 | YOLOv8n | Indian Roads (DATS) | 60.0 (Prec) | 80 |
| [17] | 2021 | YOLOv3 | Rainy Conditions | 38.5 (mAP) | 30 |
| [18] | 2025 | YOLOv8 | DriveIndia | 78.7 (mAP) | < 40 |
| Ours | 2026 | YOLOv11n | Fused (IDD + BDD) | 46.6 (mAP) | 70.9 |

Note: mAP (mean Average Precision); Prec (Precision); FPS (Frames Per Second); IDD (Indian Driving Dataset); BDD (Berkeley Deep Drive); MOD (Moving Object Detection); DATS (Dataset)

# 4 Conclusion

This study evaluated the robustness of YOLOv8n and YOLOv11n for autonomous driving by constructing a fused dataset that combines the chaotic traffic dynamics of IDD [1] with the adverse weather conditions of BDD100K [2]. Our results position YOLOv11n as a superior choice for safety-critical applications, demonstrated by a 3.2% increase in Precision over the baseline. This performance gain is driven by the architectural integration of C3k2 blocks, which enhance feature discrimination in dense, occluded scenes, thereby significantly reducing false positives and the associated risk of "ghost braking." Furthermore, comparison with state-of-the-art methods confirms that while specialized large-scale models may achieve higher absolute accuracy, YOLOv11n provides an optimal balance between accuracy and latency, reliably exceeding the 30 FPS threshold required for real-time navigation.
Regarding computational efficiency, a distinct trade-off was observed. While YOLOv8n retains a raw speed advantage (93.4 FPS vs. 70.9 FPS), YOLOv11n offers a more efficient profile with 22% fewer FLOPs, making it a theoretically more sustainable option for energy-constrained edge devices where thermal management is critical.
Despite these strengths, the study acknowledges a significant limitation in detecting the underrepresented class "Animals." The low recall rates observed in this category highlight the challenges posed by the dataset's long-tail distribution, in which the majority classes overshadow rare instances. This suggests that future architectural improvements must be coupled with targeted data rebalancing strategies to ensure the detection of rare yet potentially hazardous events.

## 5 Future Work

To enhance detection reliability, our future research will focus on overcoming data scarcity for "tail" classes through integration of Few-Shot Learning (FSL) and Zero-Shot Learning (ZSL). By leveraging vision-language models like CLIP and applying targeted augmentations such as Copy-Paste and Mosaic, we aim to adapt the model to open-world conditions and mitigate bias toward majority classes.
Furthermore, we intend to validate these algorithmic improvements through hardware-in-the-loop testing. We will deploy the optimized YOLOv11n on embedded platforms such as the NVIDIA Jetson Orin Nano and Raspberry Pi 5 to rigorously measure real-world latency and power consumption, ensuring that the theoretical computational efficiency translates into sustainable performance for autonomous vehicles.